\documentclass[letterpaper, 10 pt, conference]{ieeeconf}  

\usepackage{caption}
\usepackage{amsmath,amsfonts,mathtools}

\IEEEoverridecommandlockouts                              

\usepackage{graphicx} 
\usepackage{balance} 
\usepackage{url}

\title{\LARGE \bf iPhys: An Open Non-Contact Imaging-Based Physiological Measurement Toolbox
}

\author{Daniel McDuff$^{1}$, Ethan Blackford$^{2}$
\thanks{*This work was not supported by any organization}
\thanks{$^{1}$Daniel McDuff is at Microsoft Research,
Redmond, WA, USA
        {\tt\small damcduff@microsoft.com}}%
\thanks{$^{2}$Ethan Blackford is at Ball Aerospace, Fairborn, OH, USA
        {\tt\small eblackfo@ball.com}}%
}

\begin{document}

\twocolumn[{%
\renewcommand\twocolumn[1][]{#1}%
\maketitle
\begin{center}
    \centering
    \includegraphics[width=1\textwidth]{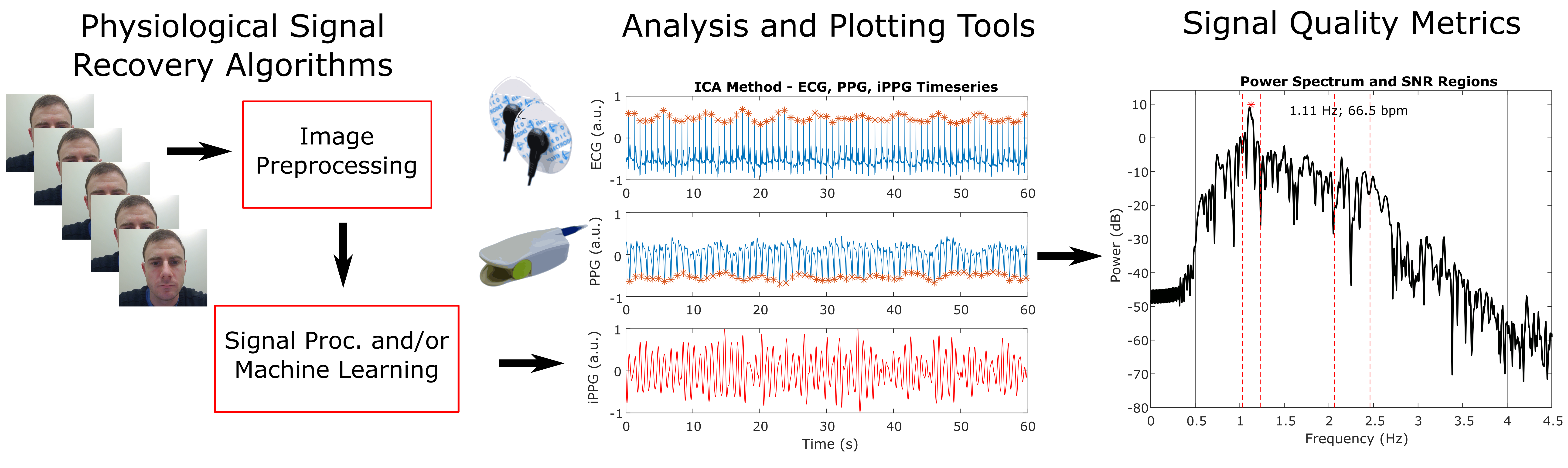} \\
    \captionof{figure}{Fig. 1. Imaging-based physiological measurement enables non-contact monitoring of cardiac and respiratory activity. We present an open toolbox that includes a range of algorithms, including those for imaging photoplethysmography and imaging ballistocardiography, and tools for plotting and evaluating signal quality.}
    \label{fig:overview}
\end{center}%
}]

\thispagestyle{empty}
\pagestyle{empty}

\begin{abstract}
Imaging-based, non-contact measurement of physiology (including imaging photoplethysmography and imaging ballistocardiography) is a  growing field of research. There are several strengths of imaging methods that make them attractive. They remove the need for uncomfortable contact sensors and can enable spatial and concomitant measurement from a single sensor. Furthermore, cameras are ubiquitous and often low-cost solutions for sensing. Open source toolboxes help accelerate the progress of research by providing a means to compare new approaches against standard implementations of the state-of-the-art. We present an open source imaging-based physiological measurement toolbox with implementations of many of the most frequently employed computational methods. We hope that this toolbox will contribute to the advancement of non-contact physiological sensing methods.
\end{abstract}

\section{INTRODUCTION}
Imaging-based methods for physiological measurement present several advantages over traditional contact-based sensing. Contact devices often require obtrusive electrodes that can become uncomfortable with extended use and can be corrupted by movement. These devices can be expensive and are not likely to be ubiquitously available. Video methods enable spatial analysis and visualization of blood flow~\cite{wu2012eulerian,hurter2017cardiolens} in addition to concomitant measurement of multiple people~\cite{poh2010non}. Cameras are ubiquitous devices and the number of imagers in the world is rapidly increasing. 

Photoplethysmography is the measurement of light reflected from, or transmitted through, the skin~\cite{allen2007photoplethysmography}. The resulting signal is known as the blood volume pulse (BVP). Imaging photoplethysmography (iPPG)~\cite{verkruysse2008remote,poh2010non,poh2011advancements} is a set of techniques for recovering these volumetric changes in blood. For a survey of iPPG techniques see~\cite{mcduff2015survey}. Ballistocardiography (BCG) uses motion of the body due to the mechanical flow of blood to measure cardiac activity~\cite{starr1939studies}. Imaging ballistocardiography (iBCG)~\cite{balakrishnan2013detecting} typically leverages optical flow estimation to track the vertical motion of the head or body from a video sequence.

Non-contact imaging-based methods have been developed to measure heart rate, respiration rate~\cite{poh2011advancements,tarassenko2014non,chen2018estimating}, heart rate variability~\cite{poh2010non,mcduff2014improvements}, blood oxygen saturation (SpO$_2$)~\cite{humphreys2007investigation,tarassenko2014non} and pulse transit time~\cite{shao2014noncontact}. Researchers have applied similar computational approaches to RGB~\cite{verkruysse2008remote,poh2010non}, IR~\cite{chen2018deepphys,chen2018estimating}, hyper- and multi-spectral~\cite{mcduff2014improvements} imager data. Measurement has been demonstrated with a range of hardware from low-cost webcams to more costly digital single lens reflex (DSLR) cameras~\cite{sun2012use} and customized devices~\cite{mcduff2014improvements}.
 
Over the past decade a community of researchers has been established around the development and application of imaging-based physiological measurement methods. Numerous computer vision and signal processing techniques have been applied in this domain to improve accuracy and robustness. Multiple workshops have been held at international conferences over the past five years and hundreds of papers have been written on the topic. However, there is a lack of publicly available tools for researchers to utilize. In other areas of biomedical engineering and computer vision the open source sharing of methods has been fundamental in advancing the state of the art~\cite{moody2001physionet,delorme2004eeglab,yan2010dparsf}. Open source methods allow researchers to compare novel approaches to consistent baselines without ambiguity about the implementation or parameters used. This transparency is important for truly evaluating how a novel approach improves the accuracy of physiological measurements.
 
PhysioNet~\cite{moody2001physionet} offers researchers access to recorded physiological signals and related open source software for analysis. This project has had a significant positive impact on the research community over the past 10 years. However, PhysioNet does not include implementations of the most commonly employed algorithms for non-contact physiological measurement. EEGLab~\cite{delorme2004eeglab} is an interactive toolbox for processing continuous and event-related electroencephalography (EEG), magnetoencephalography (MEG) and other electrophysiological data. 
DPARSF~\cite{yan2010dparsf} is a toolbox for data analysis of functional Magnetic Resonance Imaging (fMRI). All of these toolboxes were created for the MATLAB environment and have been used in a considerable number of research and teaching projects. 

Inspired by these contributions we present an open source toolbox of non-contact physiological measurement algorithms created for MATLAB. For convenience we call this the \textit{iPhys Toolbox}. This will enable researchers to present results on their datasets using standard, public implementations of the baseline methods with all parameters known and version control in place. We encourage researchers to submit new approaches to be included in this toolbox as they are published. The code and information about appropriate citations can be found at the following website. \newline \textbf{Source code:} \url{https://github.com/danmcduff/iphys-toolbox}.
 
\section{Impact of Video Parameters}
When applying imaging-based physiological measurement techniques it is important to be aware of how the nature of the data may influence the results. Researchers have characterized the impact of several parameters (imager sensor design, frame rate and resolution, video compression, ambient light, distance between the imager and the sensor) on iPPG measurement, and to a lesser extent on iBCG measurement, in a systematic manner.

Video compression can severely impact the performance of iPPG algorithms. Characterization of standard video compression algorithms has revealed that the BVP signal-to-noise ratio decreases linearly with the compression factor for the popular current
and next-generation codecs (x.264 and x.265)~\cite{mcduff2017impact}. When choosing the compression and file format for storing iPPG video datasets this must be considered.
The resolution and framerate~\cite{blackford2015effects} of a video has an expected and less significant impact on the performance of iPPG algorithms. Super resolution~\cite{mcduff2018deep} can be used to reduce the error in heart rate estimates from low-resolution videos.  Reducing the frame rate is most problematic for the estimation of peak locations when computing inter-beat intervals (IBIs) and heart rate variability metrics or pulse transit time.  It might be expected that resolution would be more problematic for iBCG approaches in which the tracking of feature points is important. However, to our knowledge this has not been systematically studied.

The use of multiple imagers can effectively boost the performance of iPPG measurement. Multiple imagers can allow for greater coverage of the skin region of interest (ROI) in the case that line of sight between one imager and skin ROI is blocked~\cite{mcduff2017fusing}. Multiple observations can also help mitigate the effect of random noise in the image sensors.

The distance range of imaging-based physiology techniques is dependant on the imaging device use. Imaging-PPG methods can be effective up to 25 meters with 4.1 BPM mean absolute error using a high-end digital camera and long distance lens~\cite{blackford2016long,blackford2016measuring}. With a low-cost webcam the performance is likely to be degraded considerably beyond a few meters.

The wavelength of the light sources also impacts the ability to recover the BVP using PPG methods~\cite{lee2013comparison}. The green light (approximately 530 nm wavelength) has been found to give superior results compared to red or blue light~\cite{verkruysse2008remote,lee2013comparison} and also compared to infra-red~\cite{maeda2008comparison}. Imaging BCG approaches are likely to be less sensitive to the spectral composition of the lighting, as long as the illumination and contrast on the faces is reasonable.

\section{Open Noncontact Imaging-based Physiological Measurement Toolbox}

The non-contact imaging-based physiological measurement toolbox contains algorithmic implementations for physiological measurement, analysis and plotting tools, and functions for signal quality evaluation (see Figure 1). For each algorithm the video frames are first analyzed to identify a region of interest (ROI). For simplicity, the ROI can be defined as the whole video frame. However, face detection is a commonly employed method for refining the ROI and tools exist for this in MATLAB. In the toolbox a line of code indicates where a custom ROI selection method could be employed. Similarly, for skin segmentation we have included flags for indicating where skin segmentation could be used and have included code for a basic rule-based skin segmentation. Below we summarize the methods, the full descriptions can be found in the referenced papers. In the code, we try to note differences from the references. In cases where filtering is used, we have tried to standardize the pass band frequencies to give the fairest comparison between methods. In our experience, filtering can have a large impact on the overall performance (whether measured via mean absolute error (MAE), root mean squared error (RMSE) or BVP signal-to-noise ratio (SNR)). This could be considered a form of over-fitting and we do not want different filtering parameters to mask the effect of improvements in the underlying signal recovery.  We describe the methods below assuming a three-channel (RGB) imager is being used. However, most methods are easily extensible to more, or fewer, channels.

\subsection{Green (Verkruysse, Svaasand \& Nelson, 2008)~\cite{verkruysse2008remote}:} 
The green channel typically contains the strongest pulsatile signal of the three RGB color channels~\cite{blackford2018remote,martinez2011optimal,verkruysse2008remote,lee2013comparison}. The first approach we have included involves spatial averaging of the green-color-channel pixels in the ROI for each frame which gives a time varying vector, x$_g$. The signal, x$_g$, is filtered using a zero-phase, 3rd-order Butterworth bandpass filter with pass band frequencies of [0.7 2.5] Hz.  

\subsection{ICA (Poh, McDuff \& Picard, 2010)~\cite{poh2010non}:} 
The Poh et al.~\cite{poh2010non} approach involves spatial averaging of pixels by color channel in the ROI for each frame to form time varing signals [\begin{math}x_r\end{math}, \begin{math}x_g\end{math}, \begin{math}x_b\end{math}]. Following this the observation signals are detrended~\cite{tarvainen2002advanced} (with $\lambda$ = 100). A z-transform is applied to each of the detrended signals. The Independent Component Analysis (ICA) (JADE implementation) is applied to the normalized color signals. The signals are filtered using a zero-phase, 3rd-order Butterworth bandpass filter with pass band frequencies of [0.7 2.5] Hz ($\sim$[42 150] BPM). A BVP source is selected by computing the Fast Fourier Transforms (FFTs) of the respective source signals and selecting the source with the highest power (normalized by total power) of the filtered signals. 

\subsection{CHROM (De Haan \& Jeanne, 2013)~\cite{de2013robust}:}
The CHROM method uses a linear combination of the chrominance signals by assuming a standardized skin-color profile to white-balance the video frames. The spatially-averaged, color-channel signals, [\begin{math}x_r\end{math}, \begin{math}x_g\end{math}, \begin{math}x_b\end{math}], are filtered using a zero-phase, 3rd-order Butterworth bandpass filter with pass band frequencies of [0.7 2.5] Hz. Following this a moving window method of length 1.6 seconds (with a step size of 0.8 seconds) is applied. Within each window the color signals are normalized by dividing by their mean value to give [\begin{math}\bar{x_r}\end{math}, \begin{math}\bar{x_g}\end{math}, \begin{math}\bar{x_b}\end{math}]. These signals are bandpass filtered using zero-phase forward and reverse 3rd-order Butterworth filters with pass band frequencies of [0.7 2.5] Hz. The filtered signals [y$_{r}$, y$_{g}$, y$_{b}$] are then used to calculate \textit{S$_{win}$}:

\begin{equation}
S_{win} = 3( 1 - \frac{\alpha}{2} )y_{r} - 2(1 + \frac{\alpha}{2})y_{g} + \frac{3\alpha}{2}y_{b}
\end{equation}

Where $\alpha$ is the ratio of the standard deviations of filtered versions of A and B:
\begin{equation}
    A = 3y_r - 2y_g
\end{equation}
\begin{equation}
B = 1.5y_r + y_g - 1.5y_b
\end{equation}
The resulting outputs are scaled using a Hanning Window and summed with the subsequent window (50\% overlap) to construct the final BVP signal.

\subsection{POS (Wang, Brinker, Stuijk \& de Haan, 2017)~\cite{Wang2016b}:}
In this method a plane orthogonal to the skin tone (POS) is used for pulse extraction. Again, the spatial average of pixels by color channel in the ROI for each frame is calculated, 
to give [\begin{math}x_r\end{math}, \begin{math}x_g\end{math}, \begin{math}x_b\end{math}]. A moving window of length 1.6 seconds (with a step size of one frame), is applied. For each window of data, the signal is divided by its mean to give [\begin{math}\bar{x_r}\end{math}, \begin{math}\bar{x_g}\end{math}, \begin{math}\bar{x_b}\end{math}]. Following this, \textit{X$_s$} and \textit{Y$_s$} are calculated where:
\begin{equation}
    X_s = \bar{x}_g - \bar{x}_b
\end{equation}
\begin{equation}
    Y_s = -2\bar{x}_r + \bar{x}_g + \bar{x}_b
\end{equation}
\textit{X$_s$} and \textit{Y$_s$} are then used to calculate \textit{S}$_{win}$, where:
\begin{equation}
    S_{win} = X_s + \frac{\sigma(X_s)}{\sigma(Y_s)}Y_s
\end{equation}
The resulting outputs of the window-based analysis are used to construct the final BVP signal in an overlap add fashion.


\subsection{BCG (Balakrishnan, Durand \& Guttag, 2013)~\cite{balakrishnan2013detecting}:}
The Balakrishnan et al.~\cite{balakrishnan2013detecting} method uses the iBCG signal to recover the HR. First, tracking of feature points using the Lucas-Kanade tracker~\cite{lucas1981iterative} is performed on the video. The y-axis components of these trajectories [\begin{math}y_1\end{math}, ...,  \begin{math}y_N\end{math}] are then filtered using a zero-phase, 3rd-order Butterworth bandpass filter with pass band frequencies of [0.7 2.5] Hz and normalized by subtracting the trajectory of the first filtered point from all others to yield [\begin{math}z_1\end{math}, ..., \begin{math}y_N\end{math}]. The overall movement vector is calculated by the square root of the summed, squared individual movements
$\sqrt{\sum_{i=1}^{N}z_i^{2}}$
and removing trajectories exceeding the 75th percentile before Principal Component Analysis (PCA) is applied. A BVP source is selected by computing the Lomb periodogram~\cite{lomb1976least,scargle1982} of the first five components. The source with the highest peak power (normalized by total power) in the range [0.7 2.5] Hz is selected.  

\subsection{Signal Quality Metrics}
We provide functions for calculating the BVP SNR ratio, mean absolute error and root mean squared error in frequency estimates.   We calculated the BVP SNR based on the approach of de Haan and Jeanne~\cite{de2013robust}. The numerator is defined as the fraction of power in the range 6 BPM either side of the gold standard pulse rate and 12 BPM either size of the first harmonic of the pulse rate. The denominator is the power in the range 0 to 240 BPM. The BVP SNR is the resulting fraction.

\section{Access and Citation}
 The \textit{iPhys Toolbox} is a set of non-contact imaging-based physiological measurement algorithms implemented in MATLAB and available for distribution to researchers online. The code and information about about the toolbox can be found at the following website: \newline \newline \textbf{Source code:} \url{https://github.com/danmcduff/iphys-toolbox}. \newline \newline The toolbox is governed by the MIT license. This allows use of the software without restriction, including without limiting rights to use, copy or modify it.  If you find this toolbox helpful or use it in research please cite this paper. Citation information can be found on the GitHub page.
 
 \section{Conclusion}
 Non-contact, imaging-based physiological measurement holds much promise and has several advantages over traditional contact sensor measurement. 
 We present an open source imaging-based, non-contact physiology toolbox which includes implementations of imaging photoplethysmography and imaging ballistocardiography methods, plotting tools and functions for calculating signal quality metrics.  We hope that this toolbox will contribute to the development of new non-contact physiological measurement methods. We encourage researchers to submit new approaches to be included in this toolbox and will continue to implement new algorithms as they are published. 
 


   
\balance{}

\bibliographystyle{ieee}
\bibliography{references}

\end{document}